# PROBABILITY HACKING AND THE DESIGN OF TRUSTWORTHY ML FOR SIGNAL PROCESSING IN C-UAS: A SCENARIO BASED METHOD

*Liisa Janssens†, Laura Middeldorp‡*

†Department of Values, Technology and Innovation, Faculty of Technology, Policy and Management, TUDelft, Delft, The Netherlands
‡Unit Defense, Safety and Security, TNO, The Netherlands

## ABSTRACT

In order to counter the various threats manifested by Unmanned Aircraft Systems (UAS) adequately, specialized Counter Unmanned Aircraft Systems (C-UAS) are required. Enhancing C-UAS with Emerging and Disruptive Technologies (EDTs) such as Artificial Intelligence (AI) can lead to more effective countermeasures. In this paper a scenario-based method is applied to C-UAS augmented with Machine Learning (ML), a subset of AI, that can enhance signal processing capabilities. Via the scenarios-based method we frame in this paper probability hacking as a challenge and identify requirements which can be implemented in existing Rule of Law mechanisms to prevent probability hacking. These requirements strengthen the trustworthiness of the C-UAS, which feed into justified trust - a key to successful Human-Autonomy Teaming, in civil and military contexts.

## 1. INTRODUCTION

The Rule of Law is a core fundament of trust in society. Mechanisms of the Rule of Law, which form the system of law, support justified trust. AI is viewed as an Emerging and Disruptive Technology (EDT) [8] and can both support and erode principles, such as the Rule of Law, that belong to constitutional societies. Activities deployed by the executive power such as by civil security authorities and the military need to adhere to the tenets of the Rule of Law. In order to counter UAS adequately, EDTs, such as AI, can become part of (new) means and methods (of warfare). Signal processing is an essential element of Counter Unmanned Aircraft Systems (C-UAS). Via a civil and military scenario, we demonstrate the potential advantages of ML for signal processing, whilst exposing vulnerabilities for maintaining the Rule of Law if probability hacking (p-hacking) is performed.

## 2. SCENARIO BASED METHOD

The operational use of ML in C-UAS can pose a challenge to the Rule of Law. Judicial consequences of design choices which can manifest during deployment are often unclear and tradeoffs cannot be overseen. Various types of requirements (technical, functional or operational) can be retrieved via the scenario-based method we developed to strengthen the Rule of Law [1, 5, 6, 7, 8, 9]. This is necessary in order to benefit from the opportunities EDTs can provide, while mitigating risks of harming the Rule of Law. For example, to mitigate risks in the use of ML, tensions between ML and the Rule of Law should be identified and translated into requirements which have institutional meaning. The scenario-based method consists of four key elements, which we will first describe and then apply to a hypothetical use case of ML for signal processing to enhance C-UAS. Furthermore, we elaborate on how p-hacking can manifest in the context of the hypothetical C-UAS enhanced with ML.

### 2.1. Capability gap analysis

The first step is a capability gap analysis: which existing problems can be solved or mitigated with EDTs? Why does this outperform current means and methods? A capability gap implies that current systems, means and methods cannot fulfil a specific capability. In C-UAS a capability gap can be defined in a kill chain, a step-by-step process deployed to defend against UAS threats. Several nations and institutions have developed their own definition of the kill chain [5, 13, 14]. In this paper, the following kill chain [11] is used:

1. Prevention: systems that collect data and information to prevent UAS threats;
2. Detection: systems indicate the presence of the UAS;
3. Tracking: if the UAS is a suspected threat, systems will track the UAS over time;
4. Classification and identification: systems classify the class of the UAS and identify the type of the UAS;

5. Designation of intent: systems that can define the intent of the UAS, i.e. what is its goal;
6. Engagement: systems that can engage the UAS by means of soft kill, e.g. jamming and spoofing, and hard kill means, e.g. kinetic means of direct energy weapons;
7. Forensics: systems that analyze the aftermath of the engagement.

For each identified capability gap, an EDT application can be proposed as a potential solution. These applications can be developed via Research & Development (R&D) or acquired by means of procurement, or a combination of both. When it comes to signal processing, ML can fill capability gaps in the detection, tracking & classification and identification step. A potential ML application could be a Recurrent Neutral Network (RNN) [15] that improves the detection of sUAS via radar. A RNN in C-UAS is evaluated via the scenario-based method in section four.

### 2.2. Scenario design

The second step is to formulate a scenario in which current means and methods cannot reach the advantage that the specific EDT can provide. The scenario describes the context in which the EDT is operationalized and contains several factors. Shaping scenarios is an ongoing iterative process. Not only traditional factors of creating scenarios in the military domain such as threat level, weather, landscape and actors, but also the legal basis and legal regime are taken into account as factors. There are four factors that need to be considered in the scenario in order to determine legitimate use of force:

1. Legitimacy: the use of force is legitimate in case of self-defense, UNSC mandate and aid to third states.
2. Type of conflict: also determined by International Humanitarian Law (IHL).
3. Environment (rural, urban, etc.).
4. Negative impact on critical infrastructure should be kept at a minimum.

The C-UAS enriched with an EDT is placed into an operational context in which there is legitimate use of force. The reason for this choice is intended to set the focus not on compliance issues whether the use of force is legitimate, but on fundamental reflections on the Rule of Law. Our research interest lies in unintentional violations of the Rule of Law once EDTs are deployed. Transparency and contestability are tenets of the Rule of Law. These can unintentionally be violated; these violations are only to be discovered via interdisciplinary analyses and therefore harder to find. In an interdisciplinary setting, unintentional violations of the Rule of Law can be identified. In this process, gaps on view points of what is at stake in the scenario are being bridged.

Once requirements are identified via the interdisciplinary scenario-based method, these can also inform compliance issues since positive law is one of the guardrails of the Rule of Law. Without this meta perspective of the lens of the Rule of Law, it is possible to get stuck in all kinds of compliance questions. For example, on the necessity of making a distinction -one of the cardinal principles of IHL- between military and civilian objects. There is no good answer to this question when it is asked at the wrong moment. When distinction is requested as a requirement in the development process of a ML application, this can form an obstacle to adherence. Therefore, the requirement that a ML application needs to prove it is able to make a distinction between civilian and military objects can hinder innovation. It can even be the case that the innovation is stopped by a legal compliance department because it is not 'compliance by design'. The question of distinction is a valid legal question, but does not have to be linked directly as a requirement to the development of ML without context. It gives a false sense of adhering to the principles of IHL. Adhering to cardinal principles takes into account more aspects than only the technical and above all: is context specific. These complexities can be captivated in a scenario and need to be studied in an interdisciplinary fashion. Identifying the capability gap as the starting point is a deliberate choice to focus on the necessity to innovate with EDTs instead of showstopping these. Compared to normal means and methods it is always needed to take into account a probability rate. The scenario-based method identifies the advantage via the capability gap analyses. Otherwise, it is not justifiable or proportional to use something that comes along with risks you cannot 'fully test off'. By identifying the advantage, the method also pushes for innovations that are solving problems, instead of just innovations for the sake of innovation.

### 2.3. Requirement identification

The third step is to analyze the scenario with the specific EDT and capability gap, linked to operational practice, via a list of requirements. In the case of AI as an EDT, there are (policy)frameworks from various entities and bodies on responsible AI. The deduction of a requirement-list depends on the desired framework for responsible AI that needs to be analyzed and contextualized via the Rule of Law together with a specific technical paradigm (for example MLOps [12]).

### 2.4. Validating, verifying and implementing requirements

Requirements retrieved from a specific technical paradigm can have institutional meaning when these are implemented in Rule of Law mechanisms, i.e. revisiting (old) legal concepts anew, policies, informing standards, R&D and procurement procedures. If there is a framework applicable for engaging with stakeholders with decision power these decision makers can be informed via the scenario. Via analysing snapshots of a scenario in the interdisciplinary setting of the scenario-based method it can become clear via deep dives which requirements are there to be found and how these strengthen the Rule of Law.

## 3. PROBABILITY HACKING AND TRANSPARANCY

P-hacking can be done by tweaking the performance metrics; applied to ML it leads to untrustworthy accuracy rates [3, 4, 6]. During military operations trustworthy ML is key to gaining a military advantage which can manifest on the strategic, tactical and/or operational level. It is likely that ML brings something new to the table that was not possible before. This aspect can be ground to justify the implementation of ML. Another ground for justification can be a limited time window to make a well-informed decision and that decision support is needed. When it comes to human oversight, generally speaking, for a commander it is cognitively impossible to determine how ML performs in real time and to review whether the outcomes of ML are trustworthy. First of all, if that would have been possible: why should you use ML instead of human cognitive capacities? With the presumption that ML can bring something different to the table than human cognitive capacities or normal means and methods entail, it can be justifiable to use such an application. The other aspect that needs to play a role is that the amount of data that is analyzed by the ML cannot be overseen by a commander. The impossibility of reviewing the outcomes is even more accentuated in the scenario of cognitive overload. Therefore, trustworthiness of the ML outcomes is crucial since a commander needs to be able to trust these blindly. In order to establish this the transparency of the R&D and/or procurement process of a ML application is important to guarantee its quality. Various requirements are needed for the design of studies to determine the performance of a system [4]. Whoever puts a ML application on the market should pre-register the research design, including the subsequent updates that were used to develop the application. This will contribute to the contestability of claims regarding the safety, security and reliability of AI applications, while also enabling the contestability of decisions based on these applications in terms of potential norm violations. If such pre-registration were to become part of R&D and/or procurement it would also be a good example of 'legal protection by design' [2, 3]. The pre-registration of research designs would create transparency regarding the processes involved in developing a ML. The details of an algorithm's history (including its development history) and performance need to be clear to who is responsible when it comes to a decision to deploy these applications. Requesting transparency of the data sets and algorithms used can prevent bias. It is important that beforehand, the choice of the data is explained, the type of error it may contain and the way it has been curated [3]. Pre-registration of the research design is required to prevent bad ML. Trust in AI should be institutionalized through interdisciplinary requirements that combine law, technology, and operational needs. ML in C-UAS can offer operational flexibility and reduced cognitive burden for commanders. But without justified trust ML in C-UAS can also lead to problems on accountability, since possibilities to align with mission intent could be made impossible.

## 4. SCENARIO-BASED METHOD: A SHOWCASE

We showcase the scenario-based method in a hypothetical use case of C-UAS enriched with ML. A concrete example of a capability gap in the detection step, is how to detect UAS with a smaller Radar Cross Section (RCS). Small UAS (sUAS) mostly fly close to the ground to extend the battery life because of less wind resistance [5]. As the sUAS is flying a low altitude, the background becomes cluttered which complicates detection of the sUAS via radar. We showcase p-hacking and the application of a RNN [15] designed to improve the radar detection of sUAS.

### 4.1. Implicit P-hacking

Imagine the RNN designed to improve the detection of sUAS in a cluttered background is developed by a promising C-UAS start-up that won a tender set out by a civil security authority. The start-up won the tender because they performed best with their performance metrics tested in both urban and rural situations. It seems that there was enough variation in both their training and validation data. Based on this performance the assumption was made that the accuracy rate was valid in all types of environments and circumstances. What is not made explicit by the start-up is that they trained the RNN on data only retrieved from Fixed Wing (FW) sUAS. During a big civil event in an urban area, a civil security officer is responsible for protecting the event against sUAS. Before the event, the officer receives notifications of a terrorist threat by use of UAS. The modus operandi of the malicious actors is with FW sUAS, and this has manifested itself earlier in another part of the country in a rural environment. Radar is applied for early detection, but since the event is in an urban area, the background is cluttered hampering timely detection. The civil security officer decides to use and trust the C-UAS radar system enhanced with the RNN, to timely detect possible sUAS. Up to now the civil security authority always faced FW sUAS challenges and the accuracy of the enhanced C-UAS system is good. However, during the event, a novelty occurs: the sUAS used by the malicious actors is a Rotary Wing (RW) sUAS. Because the RNN is trained and validated on FW data, the system fails to detect RW sUAS. The civil security officer does not see the approaching sUAS and the event is heavily disrupted. The civil security authority made the assumption that a threat always contains FW sUAS, and therefore that the enhanced C-UAS works to counter the threat. If the research design was made transparent by the start-up, this would have been prevented, because then it was made clear that the accuracy rate was not representative for RW sUAS. The start-up finetuned the model on FW data, but did not make the data versatile enough for (widely spread) RW UAS. Although this bias was introduced unintentionally, it can still be considered a form of p-hacking, as the lack of representative data skews the model's performance and evaluation outcomes.

### 4.2. Explicit P-hacking

The military scenario takes place under the framework of collective self-defence (Article 5 of the NATO Treaty). Six months earlier, a confidential procurement process was initiated to acquire a radar system capable of detecting small UAS (sUAS) with an accuracy rate of at least 85%. This threshold was agreed upon as a responsible benchmark and aligned with the idea of "compliance by design." Given the operational urgency, the military prioritized obtaining the most advanced ML application available. The manufacturer delivered a solution, a Recurrent Neural Network that met the required 85% accuracy rate. After procurement, the military requested access to the research design behind the model. However, the manufacturer refused, citing trade secrets. Prior to procurement, no consideration was given to the risk of p-hacking or measures to prevent it. For the procurer, responsible AI was equated primarily with achieving a high accuracy rate, which led to accepting the manufacturer's position without further scrutiny. Unbeknownst to the military, the RNN had been trained exclusively on data coming from ideal weather conditions: clear skies and dry terrain. In addition, an employee at the manufacturer deliberately manipulated the accuracy rate to satisfy internal expectations. In other words, the training data was tweaked so that the accuracy rate was sufficiently high. During the deployment of the enhanced C-UAS, a severe hailstorm disrupted radio signals, causing the radar system to malfunction and trigger prematurely. The commander, believing the system's output was reliable, acted immediately to counter the perceived threat. However, due to the malfunction, the response was mistimed, leaving no opportunity to reload for a second attempt. As a result, the base and its assets were destroyed. This failure was ultimately unpreventable because p-hacking, through tweaking the training and validation data such that the minimal accuracy rate was met, had compromised the integrity of the system from the outset.

### 4.3. Responsibility in the case of cumulative causality

In both showcased variations, it was not transparent that p-hacking was performed. The judicial consequences of p-hacking are real since legal responsibilities need to be addressed correctly. During operational deployment, cumulative causality can manifest, meaning it cannot be identified where an error started and which action led to what other errors. How does p-hacking affect the individual responsibility of the commander and how does this relate to state responsibility? P-hacking could have been prevented by demanding the right requirements in a pre-registration of the research design. This is not the responsibility of the civil security officer or commander, but of the government or, in NATO context, the coalition parties. The tweaked ML application was not suitable for deployment as it was untrustworthy. The accuracy rate in itself is not important; what is important is that the accuracy rate is trustworthy. The data on which the algorithm is trained should be validated for possible biases. Requirements for both variations of the scenario can be retrieved via a combination of the technical paradigm MLOps [12] and the Rule of Law tenets of transparency and contestability. For the military scenario NATO's framework for principles of Responsible Use for AI [6, 7, 8] can be added to engage with stakeholders and decision makers.

### 5. TRUST AND THE SYSTEM OF LAW

P-hacking leads to the erosion of a fair trial in front of a (military) judge, because it is not transparent to the officer or commander that p-hacking was the cause of errors he or she could be falsely accused of not doing the right thing. Trust needs to be justified trust. A false accusation at the address of an officer or a commander of not aligning with mission intent (and rules of engagement) erodes the Rule of Law tenets of transparency and contestability. Justified trust can be established when a government or coalition take their responsibilities in the value chain of partners of the R&D and/or procurement of means and methods. Commanders' intent can be aligned with mission intent and rules of engagement in a trustworthy manner if p-hacking is prevented. Transparency of research design is key in maintaining the tenet of contestability which is attributed to the Rule of Law. The scenario-based method contextualizes the advantage that is aimed for and gives the possibility to seek out the 'not low hanging fruit' requirements, i.e. which you cannot find without other disciplines. From a 'compliance by design' perspective, and without an understanding of technical paradigms, one could come up with the ideal that responsible AI can be determined in striving for an accuracy rate of (for example) 85% and higher. Pushing for high accuracy rates can lead to an incentive to do p-hacking by tweaking the performance metrics. Whilst the ideal of what responsible AI entails should be a trustworthy accuracy rate; that is not the same as the level of the outcomes. Therefore, it is important not only to ask legal questions on compliance, but also from an integrative level of understanding of what is at stake in protecting the Rule of Law. We need to innovate with EDTs to secure we have the best means and methods, whilst protecting the Rule of Law against unintentional harm via interdisciplinary informed requirements.

## 6. CONCLUSION

Trust and the system of law are complementary to each other, but cannot be replaced one for the other. A commander needs to be able to trust ML in order to legitimately engage a target. In the case of new means and methods, of which EDTs can be part, trust without formal institutionalized requirements is empty. Trust becomes justified trust when government pushed for the right requirements to enhance transparency at the right moment in the development process. P-hacking is very difficult or even impossible to discover after it has been done. Since this can also happen within the power of one individual who develops ML. Pre-registration of the research design is a good example of how setting the right requirements on transparency at the right moment can lead to justified trust. This is also coined by Hildebrandt as 'legal protection by design' [2, 3].

## 7. RELATION TO PRIOR WORK

The scenario-based method [1, 5, 6, 7, 8, 9] is developed under the thought leadership of Janssens who worked as a lawyer and philosopher amongst mathematicians, rocket scientists, computer scientists and engineers for 8 years at TNO (2017-2025). The NATO HQ study 'The Design of AI in C-UAS and the Rule of Law' (2022-2023) [7] is founded upon the chapter 'The Juridical Landscape of Countering Unmanned Aircraft Systems' [5].